# EARLY FAULT DETECTION ON CMAPSS WITH UNSUPERVISED LSTM AUTOENCODERS

**P. Sánchez, K. Reyes, B. Radu, E. Fernández**

Cátedra IA3. Dept of Computer Science, University of Alcalá, Spain
{pablo.sanchezgallego, tatiana.reyes, bianca.radu, eugenio.fernandez}@uah.es

## ABSTRACT

This paper introduces an unsupervised health-monitoring framework for turbofan engines that does not require run-to-failure labels. First, operating-condition effects in NASA CMAPSS sensor streams are removed via regression-based normalisation; then a Long Short-Term Memory (LSTM) autoencoder is trained only on the healthy portion of each trajectory. Persistent reconstruction error, estimated using an adaptive data-driven threshold, triggers real-time alerts without hand-tuned rules. Benchmark results show high recall and low false-alarm rates across multiple operating regimes, demonstrating that the method can be deployed quickly, scale to diverse fleets, and serve as a complementary early-warning layer to Remaining Useful Life models.

**Keywords:** aircraft maintenance, predictive maintenance, anomaly detection, unsupervised learning, turbofan engines

## I. INTRODUCTION

Predictive maintenance has emerged as a critical strategy for optimising the operational lifespan and reliability of complex mechanical systems. Data-driven approaches enable industries to anticipate failures, reduce downtime, and enhance maintenance scheduling through condition-based monitoring. In aerospace applications, conventional prognostic methods primarily focus on supervised learning techniques for Remaining Useful Life (RUL) estimation. These approaches require complete run-to-failure trajectories to train models that predict the remaining operational lifespan of components. However, such comprehensive failure data is often unavailable in real-world scenarios due to safety protocols and maintenance interventions that prevent equipment from reaching complete failure states, resulting in limited labelled degradation sequences for training.

The NASA CMAPSS (Commercial Modular Aero-Propulsion System Simulation) datasets (Saxena et al., 2008) have become the standard benchmark for developing and evaluating prognostic algorithms in aircraft engines. These datasets simulate realistic operational scenarios and degradation patterns across multiple engine units, providing valuable insights into failure progression. While most existing research concentrates on supervised RUL prediction methods (Fan et al., 2024; Sateesh Babu et al., 2016), the practical challenges of obtaining labelled failure data motivate the exploration of alternative approaches that do not depend on end-of-life annotations.

Beyond these well-studied supervised approaches, predictive maintenance literature also encompasses a broad family of statistical and machine-learning models that offer complementary perspectives on degradation modelling. Classical supervised methods such as linear models, logistic regression, discriminant analysis, support vector machines, and decision trees have long been applied to industrial health monitoring due to their interpretability and computational efficiency (Agresti, 2003; Anderson, 1958; Cox, 1958; Breiman, 1984; Cortes et al., 1995). More flexible ensemble techniques, including Random Forests and gradient-boosted regressors, provide improved generalisation in heterogeneous operating environments (Breiman, 2001; Friedman, 2001). In parallel, statistical degradation models such as ARIMA, path-based regression models, Wiener/Gamma stochastic processes, and Cox proportional hazards frameworks have been extensively used for remaining-life forecasting under uncertainty (Hamilton, 2020; Lu et al., 1993; Whitmore, 1995; Cox, 1972). Recent advances in deep learning have further expanded the methodological landscape through CNNs, LSTMs, autoencoders, GANs, graph neural networks, and Bayesian deep-learning architectures, enabling the extraction of complex temporal, spatial, and latent representations from multi-sensor systems (Bishop, 1995; Hinton et al., 2006; Hochreiter et al., 1997; Gui et al., 2021; Zhou et al., 2020). This diversity of techniques underscores the need for approaches that do not rely on labelled end-of-life data, particularly for safety-critical systems such as aircraft engines.

This article presents an unsupervised learning approach for early fault detection in aircraft engines using an LSTM autoencoder. The proposed system learns normal operational patterns from sensor data during healthy engine conditions, subsequently flagging deviations that indicate potential degradation. Unlike traditional RUL estimation methods, our approach specifically targets the identification of anomalous behaviour corresponding to the final 10% of engine life, providing maintenance teams with actionable alerts while requiring no labelled failure data for training. The methodology leverages the temporal dependencies inherent in engine sensor readings through the LSTM autoencoder, offering a practical solution for scenarios where complete run-to-failure data is unavailable.

## II. DATA SCIENCE METHODOLOGY

Unlike more mature disciplines such as Software Engineering, Data Science and Artificial Intelligence have historically lacked well-established methodologies, workflows, or life-cycle models that comprehensively address all phases involved in the development and deployment of AI-based solutions. It was not until the late 1980s that the first methodological proposals began to emerge, driven by the convergence of Software Engineering principles and Artificial Intelligence research. At the beginning of the 21st century, the first widely adopted methodological framework specifically oriented toward Data Mining was introduced: CRISP-DM (Cross-Industry Standard Process for Data Mining) (Shearer, 2000). This model rapidly gained recognition in both academia and industry, becoming a de facto standard for applying Artificial Intelligence techniques to decision-making and practical problem-solving.

CRISP-DM remains one of the most commonly used frameworks for data mining and data analysis across a wide range of organizations. It is structured around six iterative phases: (i) Business Understanding, where project objectives are defined in alignment with business goals; (ii) Data Understanding, which focuses on data collection and exploratory analysis; (iii) Data Preparation, involving data cleaning, integration, and transformation; (iv) Modeling, where statistical or machine learning techniques are applied; (v) Evaluation, in which model performance and alignment with business objectives are assessed; and (vi) Deployment, where the developed model is integrated into a production environment to generate actionable insights. Owing to its data-driven and iterative nature, CRISP-DM allows transitions between phases as required by intermediate results, ensuring consistency between analytical outcomes and organizational objectives. Over the years, numerous adaptations of CRISP-DM have been proposed to address diverse application contexts, including big data, cybersecurity, and fintech, among others (Rollins et al., 2015; Martínez-Plumed et al., 2021; Haakman, 2020).

The progressive integration of Artificial Intelligence with statistics, data analysis, and computer science, particularly under a big data perspective, has led to the consolidation of an interdisciplinary field commonly referred to as Data Science. In parallel, the widespread adoption of Cloud Computing platforms has enabled organizations to rapidly design, develop, and deploy Data Science projects at scale. These technological and organizational changes have fostered the emergence of more advanced methodological approaches, often promoted by major technology providers. Notable examples include the methodologies proposed by Amazon Web Services (AWS), Microsoft's Team Data Science Process (TDSP), and Google's Data Science workflows.

Beyond industry-driven proposals, a variety of methodological frameworks have been introduced and widely recognized within the Data Science community. These include, among others, CRISP-DM, CRISP-ML(Q), OSEMN, LADM, DDM, Agile Data Science, and SEMMA. From an academic standpoint, several well-defined workflows have also been proposed by researchers such as A. Tandel, J. Thomas, A. Joshi, and P. Guo, along with additional applied scientific contributions of relevance (Díaz et al., 2022; Schmetz et al., 2024; Lonescu et al., 2024; Oakes et al., 2024). This work provides a detailed review of both industrial and academic methodologies, focusing on those that exhibit sufficient structure, maturity, and practical applicability to support real-world Data Science projects.

The development of Data Science initiatives should be carried out within organizations that adopt a data-driven philosophy, combining the principles of open-standard methodologies such as CRISP-DM with agile analytics practices and project management frameworks such as PMBOK during the planning phase. This integrated approach enables implementations that are better aligned with real business needs. Furthermore, for maintenance-oriented applications, adherence to established standards such as OSA-CBM (Open System Architecture for Condition-Based Maintenance) is recommended. With respect to algorithmic model development, the use of workflows and best practices promoted by AWS Machine Learning Lens, Microsoft, and Google is encouraged, together with consolidated standards such as CRISP-DM and OSEMN, which have been refined by the Data Science community over time.

In this paper, a methodology (Moratilla et al., 2023a; Moratilla et al., 2023b) is adopted for conducting data-driven analyses in the domain of predictive maintenance. This methodology is structured around five domains comprising a total of 39 processes, as shown in Table 1. It integrates and harmonizes the main methodological proposals currently available in both industry and academia, providing a comprehensive and systematic framework suited to addressing the specific challenges associated with predictive maintenance applications.

**Table 1:** Summary of CRISP methodology.

| # | Domain | Process | Description |
|---|---|---|---|
| 1 | Business Problem | Domain Knowledge | Understanding the business context, constraints, and objectives relevant to the problem. |
| 2 | Business Problem | Data-Driven Approach | Framing the problem so that it can be addressed through data analysis and modelling. |
| 3 | Business Problem | Data Science Approach | Defining the analytical strategy, techniques, and tools to be applied. |
| 4 | Business Problem | Analytics Approach | Selecting appropriate analytical methods to extract insights and support decision-making. |
| 5 | Data Processing | Data Collection | Gathering raw data from relevant sources needed for the analysis. |
| 6 | Data Processing | Data Adequacy | Assessing whether the available data is sufficient, relevant, and representative. |
| 7 | Data Processing | Sampling | Selecting a subset of data that adequately represents the overall dataset. |
| 8 | Data Processing | Data Split (Train, Validation, Test) | Dividing data into subsets to train, validate, and test models. |
| 9 | Data Processing | Data Cleansing | Removing errors, inconsistencies, and noise from the data. |
| 10 | Data Processing | Data Balancing Analysis | Analyzing and correcting class imbalance issues in the dataset. |
| 11 | Data Processing | Exploratory Causal Analysis (ECA) | Investigating potential causal relationships among variables. |
| 12 | Data Processing | Exploratory Data Analysis (EDA) | Exploring data patterns, trends, and anomalies through statistical analysis. |
| 13 | Data Processing | Data Visualization | Representing data visually to facilitate understanding and insight generation. |
| 14 | Feature Engineering | Feature Data Transform | Applying mathematical or statistical transformations to features. |
| 15 | Feature Engineering | Feature Importance | Evaluating the relevance and contribution of each feature to the model. |
| 16 | Feature Engineering | Feature Selection | Selecting the most informative features to improve model performance. |
| 17 | Feature Engineering | Feature Extraction | Reducing dimensionality while preserving relevant information. |
| 18 | Feature Engineering | Feature Construction | Creating new features from existing data to enhance predictive power. |
| 19 | Feature Engineering | Feature Transforms | Applying polynomial or nonlinear transformations to features. |
| 20 | Feature Engineering | Feature Learning | Automatically learning feature representations from data. |

| 21 | Model Development | Model Spot Checking | Rapidly testing multiple models to identify promising candidates. |
|---|---|---|---|
| 22 | Model Development | Model Evaluation | Assessing model performance using appropriate metrics. |
| 23 | Model Development | Model Selection | Choosing the best-performing model according to predefined criteria. |
| 24 | Model Development | Model Tuning | Optimizing model hyperparameters to improve performance. |
| 25 | Model Development | Model Combination | Combining multiple models to enhance robustness and accuracy. |
| 26 | Model Development | Model Calibration | Adjusting model outputs to better reflect true probabilities. |
| 27 | Model Development | Model Uncertainty Analysis | Analyzing uncertainty in model predictions and parameters. |
| 28 | Model Development | Bias-Variance Trade-off Analysis | Evaluating the balance between model complexity and generalization. |
| 29 | Model Development | Model Interpretation | Explaining model behaviour and decision logic. |
| 30 | Model Development | Model Testing | Validating model performance on unseen data. |
| 31 | Model Development | Model Finalization | Preparing the final model for deployment. |
| 32 | Model Development | Model Saving | Storing the trained model for reuse or deployment. |
| 33 | Model Operation | Model Deployment | Integrating the model into a production environment. |
| 34 | Model Operation | Model Execution | Running the deployed model on new data. |
| 35 | Model Operation | Model Analysis | Monitoring model outputs and performance in operation. |
| 36 | Model Operation | Model Updating | Updating or retraining the model as new data becomes available. |
| 37 | AI Systems Audit | AI Framework Audit | Evaluating the overall AI framework and development process. |
| 38 | AI Systems Audit | AI Regulation Audit | Verifying compliance with applicable AI regulations and standards. |
| 39 | AI Systems Audit | AI Models Audit | Auditing models for robustness, fairness, and reliability. |

## III. ANOMALY DETECTION APPROACH

Anomaly detection offers a complementary paradigm for condition monitoring that addresses the limitations of supervised RUL prediction. This methodology operates on a fundamentally different principle: instead of attempting to estimate remaining operational time, it identifies deviations from normal operating patterns that may indicate emerging faults. The technique is particularly valuable for early fault detection when complete degradation histories are unavailable.

Within the CMAPSS framework, unsupervised anomaly-detection methods offer several distinct advantages. Modern techniques, such as deep autoencoders, one-class support vector machines (OC-SVM), isolation forests, and density-based clustering, can model healthy engine behaviour using only data from normal operation (Malhotra et al., 2016). Once trained on baseline (healthy) data, these models detect sensor readings that deviate from established patterns, potentially signalling the onset of degradation (Jakubowski et al., 2021). This paradigm is particularly valuable when early detection is more critical than precise RUL estimation and when labelled failure data are scarce (de Pater et al., 2023). Recent studies have demonstrated the feasibility of deep-learning approaches on aerospace datasets; for instance, LSTM autoencoders have successfully identified incipient faults in simulated aero-engine time-series data (Du et al., 2024). Other researchers report promising results with variational autoencoders and clustering methods for early anomaly detection in comparable prognostic problems (Jakubowski et al., 2021). Leveraging such unsupervised techniques enables timely fault detection without run-to-failure labels, complementing traditional RUL models in a practical maintenance strategy.

Furthermore, broader families of unsupervised and density-based algorithms (including k-means, Gaussian mixture models, hierarchical clustering, DBSCAN, and autoencoder-based representation learning) have shown strong performance in modelling latent structure in multivariate sensor data (Ester et al., 1996; Hastie et al., 2017). Deep generative models such as GANs and variational autoencoders further support anomaly detection by learning distributions of normal behaviour and highlighting deviations from these learned manifolds (Goodfellow et al., 2014; Sakurada et al., 2014; Akcay et al., 2018). These methodological advances collectively support the use of unsupervised architectures for health monitoring in cases where direct supervision is not feasible.

## IV.   CASE STUDY

**DATA DESCRIPTION**

The CMAPSS dataset includes turbofan engine degradation data produced using NASA CMAPSS simulation software (Frederick et al., 2007). This software models the physical behaviour of aero engines under different operating conditions. The dataset consists of degradation samples from multivariate time-series and is divided into separate training and test subsets. The time series data comes from individual engines of the same type, representing a fleet with varied operational conditions and degradation patterns.

Table 2 shows the different datasets available together with the characteristics of each one. Set FD001 represents the simplest case with a single operating condition and a single failure mode, the degradation of the HPC (High Pressure Compressor) system. On the other hand, set FD004 is the most complex, consisting of six different operating conditions and two failure modes (HPC and fan degradation).

**Table 2:** Differences between subsets of the CMAPSS use case.

| Subset | Operating conditions | Fault modes | Train engines | Test engines |
|---|---|---|---|---|
| FD001 | 1 | 1 | 100 | 100 |
| FD002 | 6 | 1 | 260 | 259 |
| FD003 | 1 | 2 | 100 | 100 |
| FD004 | 6 | 2 | 249 | 248 |

Each observation in the CMAPSS dataset consists of a timestamped vector of variables, including the engine unit identifier, the operational cycle, three operational condition settings, and 21 sensor measurements. These sensor readings capture various physical properties of the engine such as temperatures, pressures, and rotational speeds at different internal components.

**DATA PREPARATION**

Before applying any preprocessing techniques, it is essential to partition the dataset to distinguish between the normal operation phase (healthy state) and the degraded operation phase. In this project, we adopt a simple yet effective strategy: the first 85% of each time series is used as the training set. This decision is based on the structure of the CMAPSS dataset, where the initial portion of each engine's life corresponds to healthy behaviour, while degradation typically begins in the later stages. Reserving the final 15% of each unit's timeline allows us to focus model evaluation on the period where faults are likely to emerge.

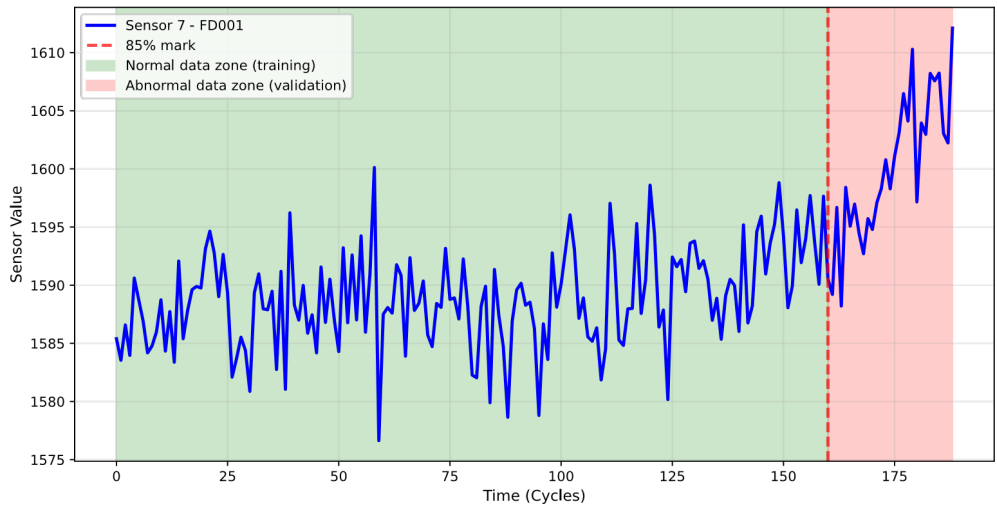

**Figure 1:** Example of partitioning on sensor #7 set FD001.

This partitioning is illustrated in Figure 1. The green-shaded region represents the training set, consisting of data from the normal operation phase. The red-shaded region, on the other hand, denotes the validation set that captures the onset and progression of degradation. The vertical red dashed line marks the 85% cutoff point, clearly separating the two phases. This split includes a 5% buffer before the last 10% of RUL, ensuring a clear boundary between healthy and degraded conditions.

This separation serves three key purposes. First, it ensures that all models are trained exclusively on data representing normal engine operation, establishing a clear baseline of healthy behaviour. Second, it prevents any degradation patterns from contaminating the training process, which is crucial for effective anomaly detection. Third, this methodology reflects real-world operational scenarios where systems are typically monitored during normal operation before being deployed to detect emerging faults.

By reserving the final 15% of each time series for validation and testing, we simulate realistic conditions for fault detection. The model's performance is evaluated on its ability to detect deviations as the system transitions into its failure phase, particularly within the final 10% of the unit's lifespan, when accurate predictions are most critical. The added 5% buffer before this phase helps strengthen the separation, allowing models to generalise better and reducing the risk of data leakage from early degradation patterns into the training phase.

The CMAPSS datasets FD002 and FD004 present a unique preprocessing challenge due to their inclusion of multiple operational conditions, reflected in three continuous operational variables. These varying conditions significantly influence sensor readings, making direct analysis of raw sensor data inappropriate for degradation modelling. Traditional approaches often cluster operational conditions into discrete states represented by binary indicators. However, we adopt a more sophisticated normalisation approach that explicitly removes operational condition effects while preserving degradation-related patterns.

Our preprocessing methodology follows a regression-based normalisation procedure similar to that of Wang et al. (2018, 2019). For each of the 21 sensor signals, we train a Multilayer Perceptron (MLP) regression model that predicts expected sensor values ($y_{pred}$) given the three operational condition variables. The normalisation is then performed by subtracting the condition-dependent predictions from the actual sensor readings:

$$x_{norm} = x_{raw} - y_{pred}, \quad (1)$$

where $x_{raw}$ represents the original sensor measurement and $x_{norm}$ is the condition-normalised value. This model, trained exclusively on healthy operation data from the training set, learns the conditional relationship between operational parameters and sensor behaviour under normal conditions.

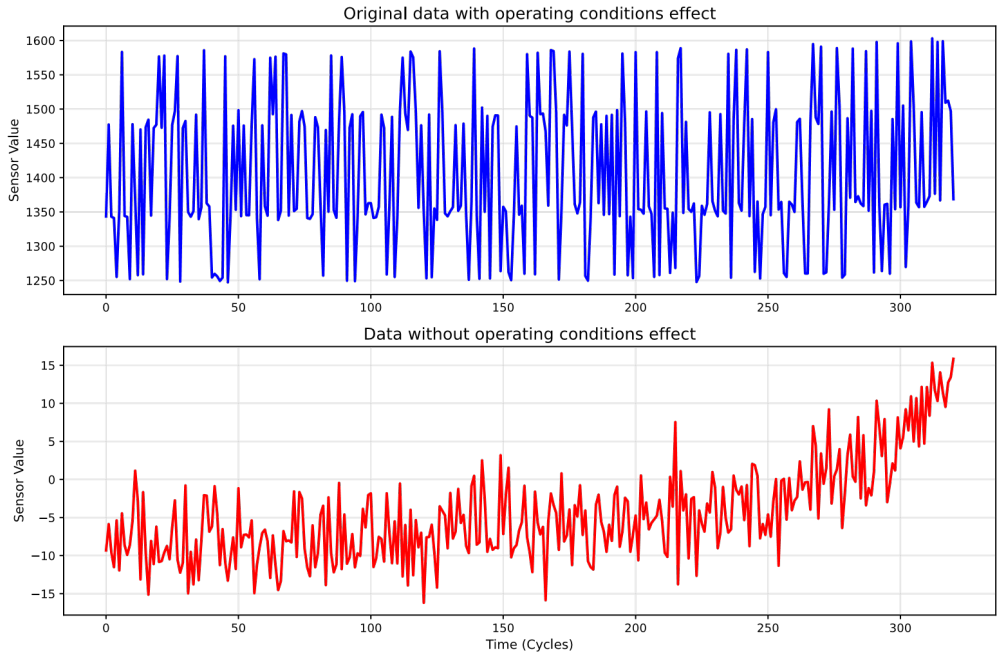

**Figure 2:** Removing the effect of operating conditions on sensor #7.

As demonstrated in prior work and visualised in Figure 2, this normalisation reveals hidden trends in sensor data that are otherwise obscured by operational variability. The resulting normalised sensor values better reflect the true health state of each engine, providing cleaner inputs for subsequent autoencoder training and anomaly detection. For FD001 and FD003 datasets (single operational condition), we apply conventional min-max normalisation, as operational condition effects are not present. The normalisation is applied as follows:

$$x_{norm} = \frac{x - x_{min}}{x_{max} - x_{min}}. \qquad (2)$$

This preprocessing strategy is particularly crucial for our unsupervised approach, as it ensures that the autoencoder learns to reconstruct and detect anomalies in condition-independent degradation patterns rather than operational artefacts. The effectiveness of this method has been validated in previous RUL estimation studies and proves equally valuable for anomaly detection tasks.

**EXPLORATORY DATA ANALYSIS**

The initial exploratory analysis of the CMAPSS dataset reveals several important characteristics of the sensor data. While some sensors show clear temporal patterns related to engine degradation, others appear relatively constant or exhibit minimal variation throughout the operational life of the engines. Traditional supervised approaches might consider eliminating these apparently uninformative sensors to reduce dimensionality and noise. However, our unsupervised approach, based on autoencoders, takes a fundamentally different perspective.

From the correlation analysis (Figure 3), we observe that several sensor pairs show high collinearity during normal operation, while some sensors demonstrate near-zero variance. Notably, sensors 1, 5, 10, 16, 18 and 19 exhibit particularly low variability across all operational conditions. In a supervised learning context, these would typically be excluded as non-informative features. However, in our unsupervised framework, we intentionally retain all sensors for three key reasons.

First, the autoencoder's latent space representation can automatically learn to weight sensor importance during training, effectively performing implicit feature selection. The network architecture naturally compresses the input dimensions while preserving the most relevant patterns for reconstruction. Second, while some sensors may appear uninformative during normal operation, their behaviour during fault conditions might contain subtle but critical anomalies that would be missed if excluded a priori. Third, the interaction between multiple "low-variance" sensors might collectively contain valuable information that individual sensors lack.

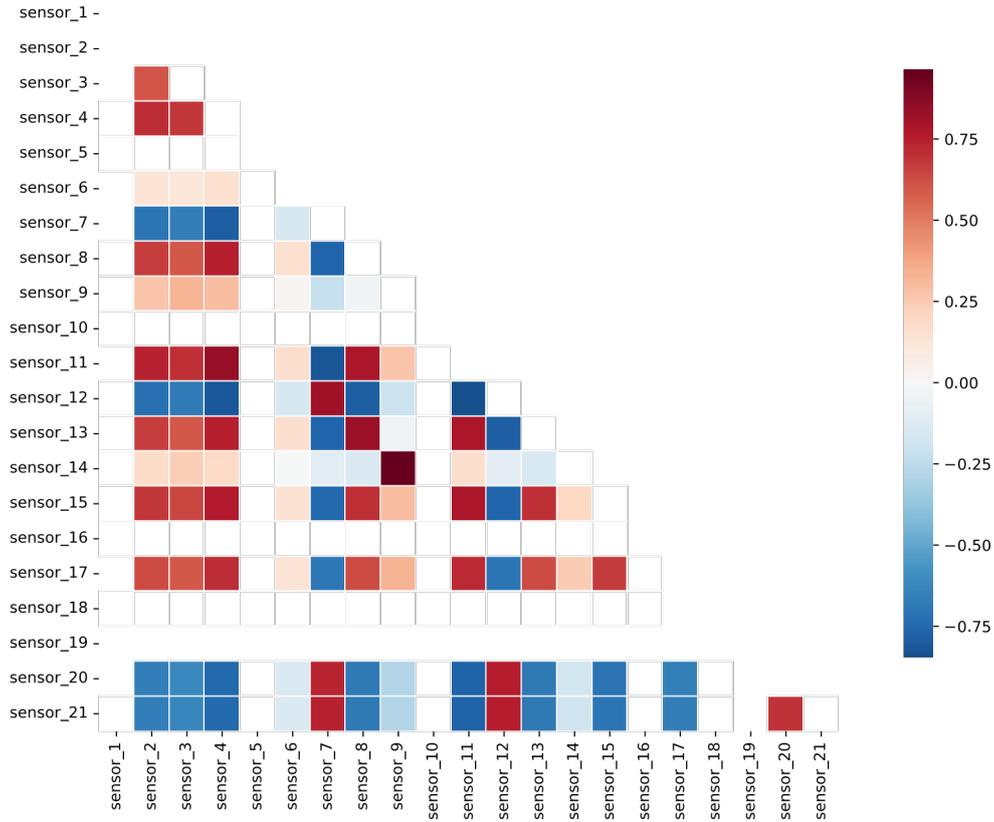

**Figure 3:** Correlation matrix of sensor readings during normal operation phase.

The temporal analysis (Figure 4) shows that most sensors maintain stable baselines during normal operation, with deviations becoming apparent only during the degradation phase. This reinforces our decision to train exclusively on the first 85% of each time series, as it provides the clearest representation of healthy operation. Interestingly, some sensors that appear constant during normal operation develop clear degradation signatures later in the life cycle, further validating our decision to retain all sensors.

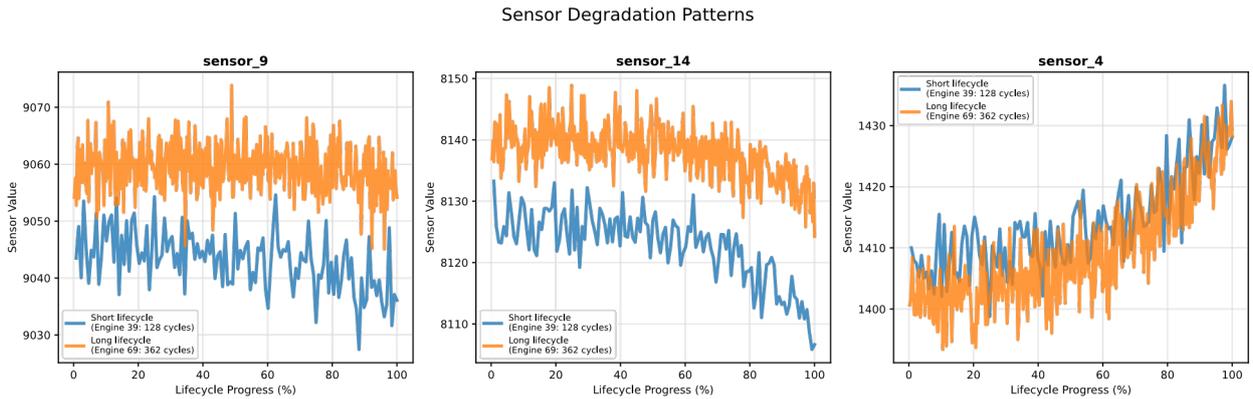

**Figure 4:** Temporal evolution of selected sensor readings showing normal operation and degradation phases.

The operational condition analysis reveals that sensor behaviours cluster distinctly under different operating regimes, as shown in the original data with operating conditions effect (Figure 2). This finding motivated our operational condition normalisation approach described in Section 3.3, as it confirms that raw sensor values are heavily influenced by external operating parameters rather than solely reflecting the engine health state.

This comprehensive EDA supports our methodological decisions to: (1) retain all sensors despite apparent low variability in normal operation, (2) employ an autoencoder architecture capable of learning meaningful latent representations, and (3) implement operational condition normalisation before anomaly detection. The analysis confirms that our unsupervised approach can potentially capture subtle degradation patterns that traditional feature selection methods might overlook during manual preprocessing (Jakubowski et al., 2021).

**FEATURE ENGINEERING**

Customary feature engineering in predictive maintenance commonly depends on manually selecting and extracting handcrafted features using raw sensor data like statistical moments (mean, variance), time-domain characteristics, or frequency-domain decompositions. However, fault detection's most unique patterns might not always be captured through these methods, especially in aircraft engines and complex systems (Ramasso et al., 2014).

In this work, we adopt an alternative model where the model itself performs feature extraction automatically during training, using the latent space representation for this purpose. When trained, a deep LSTM autoencoder neural network learns this representation to reconstruct its input data through a compressed bottleneck layer. To preserve the most salient features for reconstruction, the encoder component learns to project high-dimensional sensor readings into a lower-dimensional latent space. Nonlinear relationships and temporal dependencies in the data are intrinsically captured by this latent representation without explicit feature engineering.

**MODEL**

*Model architecture.*

The proposed anomaly detection system employs an LSTM autoencoder, a specialized neural network architecture particularly suited for capturing temporal dependencies in multivariate time series data. The model consists of two fundamental components:

- Encoder: A stack of LSTM layers that processes an input sequence of sensor readings and compresses it into a fixed-length latent vector. At each time step $t$, each LSTM cell updates its hidden state $h_t$ and cell state $C_t$ using gated operations (forget, input, and output gates) on the current input $x_t$ and the previous hidden state $h_{t-1}$. The final hidden state at the end of the sequence becomes the latent representation (the bottleneck feature vector), capturing the information of the entire window of input data.

- Decoder: A corresponding stack of LSTM layers that takes the latent vector and attempts to reconstruct the original sequence of sensor readings. This decoder essentially learns to generate the expected sensor values at each time step given the encoded summary of the sequence.

For the CMAPSS case study, we implemented a 3-layer LSTM encoder with hidden state dimensions of 16, 8, and 4, respectively. The last encoder layer's output is a 4-dimensional latent vector (bottleneck). The decoder is a mirror of the encoder in architecture, using LSTM layers to expand the 4-dimensional code back to the full sequence dimension. The relatively small size of the latent space (4) forces the network to learn a compact representation of normal engine behaviour, which helps it generalise and detect anomalies as reconstruction errors when patterns deviate from this normal manifold. Similar LSTM-based encoder–decoder architectures have been applied successfully to anomaly detection tasks in other domains, demonstrating the ability of a deep compressed model to learn complex temporal patterns of normal operation.

*Problem-specific adaptation.*

The model is specifically adapted for aircraft engine anomaly detection through three key design choices:

1. Temporal Window Processing: Input sequences are structured as sliding windows of 10 operational cycles (approximately 5% of average engine life, depending on the specific dataset), allowing the model to learn both instantaneous sensor relationships and their temporal evolution.

2. Condition-Aware Training: The preprocessed sensor data (after operational condition normalisation) is augmented with the three operational condition variables as auxiliary inputs, enabling the model to implicitly learn condition-specific normal patterns.

*Training protocol.*

The LSTM autoencoder is trained exclusively on healthy data (the first 85% of the cycles for every engine), and the remaining 15% is held out solely for post-training evaluation of anomaly detection. All windows extracted from the healthy portion of every engine constitute the training set. The network minimizes the mean-squared reconstruction error (MSE) using the Adam optimiser with an initial learning rate of $1 \times 10^{-3}$. Twenty percent of the healthy windows are randomly set aside as an internal validation split. We employ early stopping: training halts if the validation MSE does not improve for 10 consecutive epochs, preventing overfitting to noise in the baseline behaviour.

Throughout training, care was taken to prevent any information from the test (degradation) phase from leaking into the model training. All hyperparameters and threshold selections were made using only training-phase data and validation splits from that training data, mimicking a realistic scenario where one would train on past fleet data and then deploy the model on new engines.

## V. EVALUATION

**ANOMALY DETECTION MECHANISM**

During operation, the trained LSTM autoencoder continuously ingests sliding windows of raw sensor measurements and outputs a reconstructed counterpart for each window. Let $x_t \in R^{s \times w}$ denote the stack of $s$ sensor signals sampled over $w$ time steps at window index $t$, and let $\hat{x}_t$ be its reconstruction. The discrepancy between the two is quantified by the window-level mean–squared error, which serves directly as the anomaly score.

$$MSE^t = \frac{1}{sw}\sum_{i=1}^{s} \sum_{j=1}^{w}\left(x_{t,i,j} - \hat{x}_{t,i,j}\right)^2. \quad (3)$$

To convert this score into a binary decision, we calibrate a threshold using the training set exclusively. From the empirical distribution of their reconstruction errors, we compute the mean $\mu_{MSE}$ and standard deviation $\sigma_{MSE}$. The decision boundary is then fixed as

$$\tau = \mu_{MSE} + \lambda \sigma_{MSE}, \quad (4)$$

where the multiplier $\lambda$ (e.g., $\lambda = 2.5$) is a tunable hyperparameter. During deployment, any sequence whose reconstruction error exceeds $\tau$ is flagged as anomalous. Adjusting $\lambda$ allows us to trade off between false-alarm rate and missed detections, while anchoring the threshold to the observed variability of genuinely healthy data. Figure 5 shows the distribution of the errors in the different sets of the FD001 datasets.

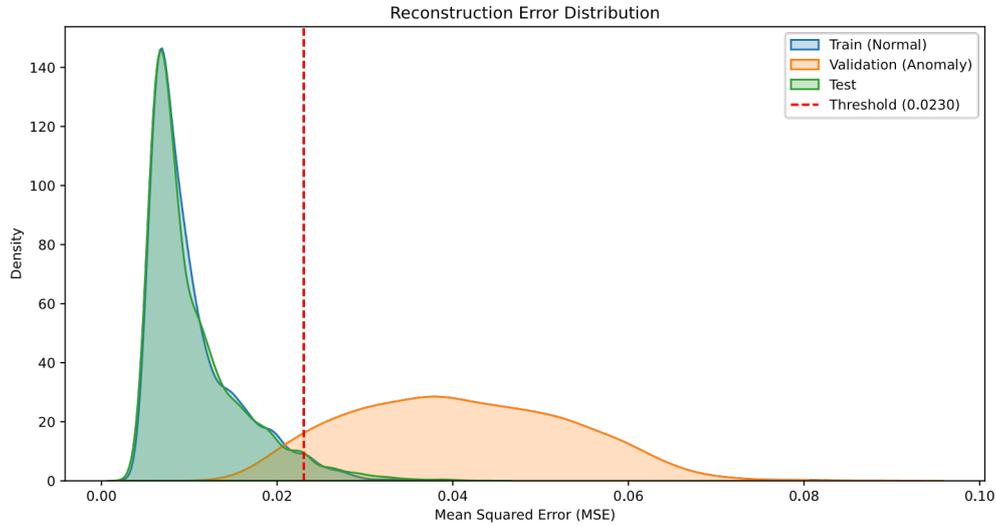

**Figure 5:** Error distribution - FD001.

**EVALUATION PROTOCOL**

Because the original CMAPSS files do not annotate the exact onset of degradation, we impose an internal partition: the first 90% of cycles in each trajectory are treated as normal, while the remaining 10% are labelled degraded. After processing a complete trajectory, the detector's predictions are cross-tabulated against these in-house labels to build the confusion-matrix counts {*True Positives (TP), False Positives (FP), True Negatives (TN), False Negatives (FN)*}. Precision, recall (sensitivity), specificity and $F_1$-score are then derived, measuring simultaneously (i) how reliably the system raises an alert during the critical last 10% of life and (ii) how rarely it emits false alarms during the much longer healthy phase. Together with early-warning-time statistics, these metrics form the quantitative basis of the results discussed in the following section.

# VI. RESULTS

Figure 6 illustrates the end-to-end detection pipeline for a representative engine (unit 81 in subset FD001). The blue curve shows the reconstruction error, i.e. the MSE between the input window and the LSTM autoencoder reconstruction, computed for each operational cycle. During nominal operation (cycles 1–180), the error fluctuates within a narrow band well below the data-driven anomaly threshold. From cycle ≈185 onward the error begins to drift upward, crossing the threshold at cycle 195, where the point that corresponds to 90% of the engine's life (orange dotted line). Every window whose error exceeds the threshold is flagged as an *anomaly* (red markers).

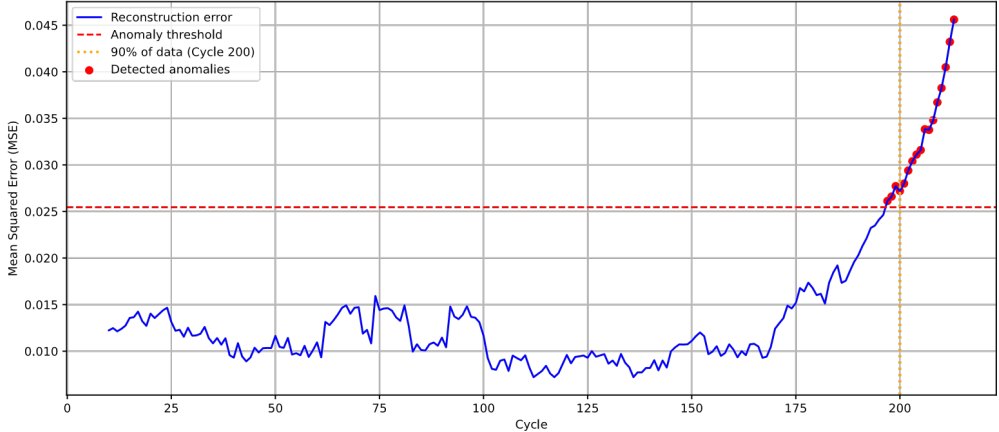

**Figure 6:** Reconstruction Error per cycle - Engine 81 - FD001.

Table 3 summarises the detection performance of the proposed LSTM autoencoder model on the four CMAPSS subsets. Although the class imbalance is extreme (between 0.8% and 1.3% of the windows in each subset are labelled as anomalous), the autoencoder achieves very high recall in all cases (≥ 0.725) while keeping specificity above 0.96, indicating a low false-alarm rate during healthy operation.

*Single-regime subsets (FD001, FD003)*

For FD001 and FD003—where only one operating regime is present—the model attains perfect or near-perfect recall (0.981 and 1.000, respectively). Precision, however, remains modest (0.196 and 0.241), pulling the $F_1$-scores down to 0.327 and 0.388. This behaviour is expected in highly imbalanced settings: a detector tuned for maximum sensitivity will inevitably flag some healthy windows as suspicious, reducing precision (Malhotra et al., 2016).

While the precision scores reported (approx. 0.20–0.37) indicate a substantial rate of false alarms, this trade-off is deliberate and aligned with the operational requirements of aerospace maintenance. In safety-critical systems, the cost function is highly asymmetric: a missed detection (False Negative) presents a catastrophic risk to flight safety, whereas a false alarm (False Positive) incurs a manageable operational cost associated with unscheduled inspections. Thus, this trade-off is acceptable and consistent with the figures reported in previous unsupervised work on CMAPSS (de Pater et al., 2023).

Furthermore, the proposed unsupervised framework serves as a first-stage screening tool within a broader maintenance workflow. Its primary design goal is high sensitivity (Recall ≥ 0.98 in single-regime datasets) to ensure that no potential fault goes unnoticed. The alerts generated by the autoencoder are intended to trigger expert review or secondary, fine-grained diagnostics, rather than immediate autonomous engine shutdown. From this perspective, maximizing Recall is the priority, establishing the model as a robust safety net.

*Multi-regime subsets (FD002, FD004)*

FD002 and FD004 pose a substantially harder problem: six distinct operating conditions interact with the sensor dynamics, producing a broader range of normal patterns. Even after the regression-based normalisation, residual regime effects can shift sensor baselines enough to mimic fault signatures, complicating the separation between healthy and anomalous windows (Baptista et al., 2021). The model still maintains good specificity (≥ 0.982), but recall drops to 0.799 (FD002) and 0.725 (FD004), and the overall $F_1$-score falls to 0.510 and 0.456, respectively. Nevertheless, recall in these multi-regime sets still exceeds the 0.70 benchmark often cited for practical early-warning systems (Yan et al., 2023), while the high specificity ensures that maintenance teams are not overwhelmed by false alarms.

*Discussion of metric interplay*

Because precision and recall respond oppositely to the decision threshold, the reported $F_1$-scores reflect a conscious bias towards high recall, which is a common choice in prognostics where missing an incipient fault is costlier than an unnecessary inspection. If required, the threshold could be tightened to boost precision (e.g., for on-wing monitoring with limited maintenance slots), at the expense of detecting some faults slightly later. Finally, we note that the extreme imbalance also deflates the apparent $F_1$: despite a score of 0.327 on FD001, the detector correctly flags 98% of true anomalies while raising false alarms in only 3.4% of healthy windows.

**Table 3:** Detection performance on the four CMAPSS subsets.

| Dataset | Precision | Recall (Sensitivity) | Specificity | $F_1$-score | Anomaly% |
|---|---|---|---|---|---|
| FD001 | 0.196 | 0.981 | 0.966 | 0.327 | 0.8% |
| FD002 | 0.374 | 0.799 | 0.982 | 0.510 | 1.3% |
| FD003 | 0.241 | 1.0 | 0.963 | 0.388 | 1.2% |
| FD004 | 0.336 | 0.725 | 0.985 | 0.456 | 1.1% |

## VII. CONCLUSIONS AND FUTURE WORK

*Limitations and future work.*

Although the proposed LSTM autoencoder delivers high recall with few false alarms, several caveats limit immediate industrial deployment. First, the 85/15 normal–degraded partition is only a heuristic proxy for real operations; engines may transition to fault states at different points, so a principled regime-change detector or Bayesian change-point model is needed to learn boundaries automatically.

Sudden load changes or sensor glitches can produce short-lived false positives. To improve robustness we already propose a *temporal persistence requirement*: an alert is raised only when the reconstruction error exceeds the threshold for at least $k$ consecutive windows (with $k$ =5, which represents ≈2% of an engine trajectory). Future work must validate the optimal choice of $k$ under different duty cycles and cost models.

Even after regression-based normalisation, residual operating-condition effects remain the dominant source of error in the multi-regime datasets FD002 and FD004. Domain-adversarial training or contrastive representation learning could help produce condition-invariant embeddings that generalise better across regimes.

*Conclusion.*

This work demonstrates that an entirely unsupervised LSTM autoencoder, trained only on baseline data and preceded by a regression-based operating-condition normalisation, can serve as a practical early-fault detector for turbofan engines. By learning the manifold of healthy behaviour and measuring reconstruction error online, the model flags deviations well before traditional RUL estimates would approach zero, even though no failure labels are required for training.

Three core contributions underpin this outcome. First, the regression-based normalisation successfully removes multi-regime artefacts while preserving subtle degradation trends, enabling a single detector to operate across heterogeneous duty cycles. Second, a compact four-dimensional latent representation captures essential engine dynamics, allowing robust anomaly detection with modest computational overhead. Third, the evaluation protocol (training on the first 85% of life and testing on the final 10%) closely mirrors industrial practice and shows that the method achieves high recall with a low rate of spurious alerts.

Beyond these technical advances, the framework complements existing RUL models by providing an earlier, confidence-weighted indicator of incipient faults, thereby supporting condition-based maintenance decisions without increasing the false-positive burden. Although further work is needed to adapt thresholds dynamically, enhance interpretability, and bridge the simulator-to-flight gap, the present study offers a label-free foundation on which hybrid, data-plus-physics strategies for safety-critical prognostics can be built.

## ACKNOWLEDGEMENTS

This work has been carried out within the framework of the "Cátedra ENIA IA3: Cátedra de Inteligencia Artificial en Aeronáutica y Aeroespacio", subsidized by the "Ministerio de Asuntos Económicos y Transformación Digital (Secretaría de Estado de Digitalización e Inteligencia Artificial), del Gobierno de España.